\documentclass[conference,a4paper]{IEEEtran}
\IEEEoverridecommandlockouts
\usepackage{cite}
\usepackage{amsmath,amssymb,amsfonts}
\usepackage{algorithmic}
\usepackage{graphicx}
\usepackage{textcomp}
\usepackage{xcolor}
\usepackage{tabularx}
\usepackage{booktabs}

\def\BibTeX{{\rm B\kern-.05em{\sc i\kern-.025em b}\kern-.08em
    T\kern-.1667em\lower.7ex\hbox{E}\kern-.125emX}}
\begin{document}

\title{KENGIC: KEyword-driven and N-Gram Graph based Image Captioning
\thanks{\copyright2022 IEEE. Published in the Digital Image Computing: Techniques and Applications, 2022 (DICTA 2022), 30 November – 2 December 2022 in Sydney, Australia. Personal use of this material is permitted. However, permission to reprint/republish this material for advertising or promotional purposes or for creating new collective works for resale or redistribution to servers or lists, or to reuse any copyrighted component of this work in other works, must be obtained from the IEEE. Contact: Manager, Copyrights and Permissions / IEEE Service Center / 445 Hoes Lane / P.O. Box 1331 / Piscataway, NJ 08855-1331, USA. Telephone: + Intl. 908-562-3966.}}

 \author{\IEEEauthorblockN{Brandon Birmingham and Adrian Muscat}
 \IEEEauthorblockA{\textit{Department of Communications and Computer 
 Engineering} \\
 \textit{University of Malta}, Msida, Malta \\
 \{brandon.birmingham.12, adrian.muscat\}@um.edu.mt}}


\maketitle

\begin{abstract}
This paper presents a Keyword-driven and N-gram Graph based approach for Image 
Captioning (KENGIC). Most current state-of-the-art image caption generators are 
trained end-to-end on large scale paired image-caption datasets which are very 
laborious and expensive to collect. Such models are limited in terms of their 
explainability and their applicability across different domains. To address 
these limitations, a simple model based on N-Gram graphs which does not require 
any end-to-end training on paired image captions is proposed. Starting with a 
set of image keywords considered as nodes, the generator is designed to form a 
directed graph by connecting these nodes through overlapping n-grams as found 
in a given text corpus. The model then infers the caption by maximising the 
most probable n-gram sequences from the constructed graph. To analyse the use 
and choice of keywords in context of this approach, this study analysed the 
generation of image captions based on (a) keywords extracted from gold standard 
captions and (b) from automatically detected keywords. Both quantitative and 
qualitative analyses demonstrated the effectiveness of KENGIC. The performance 
achieved is very close to that of current state-of-the-art image caption 
generators that are trained in the unpaired setting. The analysis of this 
approach could also shed light on the generation process behind current top 
performing caption generators trained in the paired setting, and in addition, 
provide insights on the limitations of the current most widely used evaluation 
metrics in automatic image captioning.
\end{abstract}


\begin{figure*}[t]
	\begin{center}
		\includegraphics[width=0.7\linewidth]{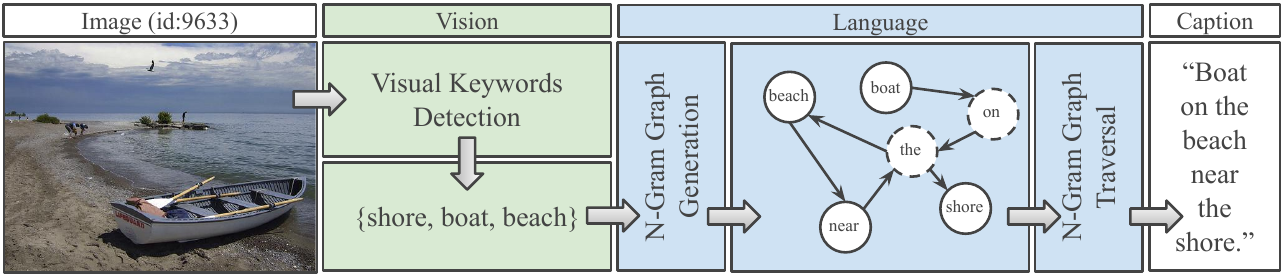}
	\end{center}
	\caption{High-level architecture of KENGIC.}
	\label{fig:kengic}
\end{figure*}

\section{Introduction}
Image captioning is the process of automatically describing images with 
succinct and human-like sentences. In the last decade, this research area 
has gained huge popularity in both academia and industrial key players 
within the Artificial Intelligence (AI) domain. This research area finds 
application (a) in content-based image retrieval, (b) in enhancing the 
accessibility 
of image collections, and (c) as an aid for the visually impaired to 
better help them understand the visual world through spoken feedback.
Despite its advantages, one has to be aware that such technology could reduce 
people's privacy, opens the door to impermissible analysis of personal image 
repositories and its output could disadvantage groups not well represented in 
the datasets. In addition, cases where visual content is incorrectly described 
can lead to dangerous predicaments in critical scenarios (e.g., a visually 
impaired person crossing a road). It is therefore important that prior to any 
deployment,
such systems are thoroughly tested against these characteristics 
and its limits in accuracy are known.

Although significant progress has been made with the advancement that Deep 
Learning (DL) 
has brought within the AI community, modelling the intersection between Vision 
and Language still remains one of the current challenging issues in AI. 
State-of-the-art image caption generators are generally trained end-to-end on 
large image-caption pair datasets using an encoder-decoder 
architecture~\cite{chen2015mind}. These are trained to 
simultaneously encode images into visual features and decode these learned 
embeddings sequentially into word sequences.

Recently, the Transformer~\cite{vaswani2017attention} model which showed to be 
very 
promising in machine translation has achieved state-of-the-art 
results in image caption generation. This model eschews the 
recurrence of conventional decoders and instead relies on an attention 
mechanism~\cite{xu2015show} to draw global dependencies 
between the vision and language 
modalities. Although encoder-decoder based models outperformed previous 
template~\cite{farhadi2010every} and 
retrieval~\cite{ordonez2011Im2Text} based models, they 
lack in compositional generalisation and are 
prone to generate biased~\cite{hendricks2018women} and non-grounded 
captions~\cite{zhou2020more} since they do not explicitly detect objects and 
their 
corresponding relationships. To mitigate these problems, scene 
graphs~\cite{xu2018scene} have 
recently been used to encode the structural representation of images. Recently, 
researchers started focusing on the long-term viability of image caption 
generators by reducing the dependency of image and caption paired data and 
proposed models that are trained in an unpaired setting where no association 
between images and captions is made during the training 
phase~\cite{cao2020interactions,ben2022unpaired}. Unsupervised image 
captioning~\cite{feng2019unsupervised} is also attracting attention to further 
reduce the dependency of having text corpora related to images. To 
address and perhaps understand the underpinning limitations in image caption 
generation, this paper proposes a keyword driven and $n$-gram 
graph-based image captioning (KENGIC) approach. This was purposely developed to 
(a) reduce the dependency of paired image and caption datasets, (b) provide 
explainable and traceable caption generation, (c) investigate the role of 
visual keywords in image captioning whilst projecting insights on how captions 
can be generated from a set of relevant image keywords without using any 
end-to-end learning unlike current state-of-the-art image caption generators.
Therefore, the main contributions of this paper are twofold: (a) it proposes a 
novel image 
caption generator based on a keyword-driven and $n$-gram graph based approach 
and (b) presents a detailed study on the use of both human authored and machine 
generated keywords in this approach. 
The latter study investigates the sets of keywords that result in higher 
evaluation scores and therefore provides insights 
into the automatic metrics that are commonly used to evaluate captions and 
optimise image captioning models.

\section{Related Work}
In the last decade, the task of automatic image caption generation attracted a 
lot of attention from both Computer Vision (CV) and Natural Language Processing 
(NLP)
researchers. Early contributions cast the task of image captioning as a 
retrieval- or summarisation-based problem. Models were developed to reuse and 
synthesise captions from already constructed collections of image-sentence 
pairs~\cite{ordonez2011Im2Text} or from associated text 
documents~\cite{aker2010generating}.
Retrieval-based models were proposed as an attempt to retrieve human-like 
captions by ranking captions from visual 
space~\cite{ordonez2011Im2Text}
or from multimodal 
space~\cite{jia2011learning}
which combines both textual and visual data in one bi-directional common 
space. 

Since being dependent on similar images often tends to be less effective in 
handling the complexities and object combinations of natural images, early 
researchers have also opted for a 
direct generation pipeline to generate novel 
captions~~\cite{farhadi2010every,li2011composing}. This 
approach, which is the one mostly related to this work, first uses visual 
detectors to  obtain an explicit intermediate meaningful 
representation of the image, and secondly, the generation phase turns the set 
of 
detected visual keywords into captions by using natural language generation 
techniques, such as templates, $n$-gram models and grammar rules. For instance, 
Farhadi et al.,~\cite{farhadi2010every} used detections to construct triplets 
of $\langle object,action,scene\rangle$ which were then translated to 
text using a template. Similarly, Li et al.~\cite{li2011composing} generated 
descriptions based on computer vision inputs such as objects, visual attributes 
and spatial  
relationships using web-scale $n$-grams. Yang et al.,~\cite{yang2011corpus} 
constructed captions by filling templates with the most likely objects, 
verbs, prepositions and scene types based on a Hidden Markov 
Model. Kulkarni et al.,~\cite{kulkarni2011baby} detected image objects, 
attibutes and 
prepositional relationships and used a Conditional Random Field model to 
incorporate the unary image potentials with higher order text-based potentials 
as computed from large textual corpora. Image captions were then generated 
based on the predicted graph labels. To encode the geometrical relations 
between image regions, Elliott and Keller~\cite{elliott2013image} proposed the 
visual dependency 
representations (VDRs). This intermediary representation was traversed to 
insert keywords in 
sentence templates by their proposed visual dependency grammar. Other image 
captioning models were based on more complex language models. For 
example, Mitchell et al.,~\cite{mitchell2012midge} implemented an 
overgenerate-and-select 
approach-based system 
by which syntactically correct sentence fragments based on detected visual 
keywords were generated and combined via a tree-substitution 
grammar. A more flexible approach was later proposed by Kuznetsova et 
al.~\cite{kuznetsova2014treetalk}. In the latter work, a stochastic tree 
composition algorithm designed to combine tree fragments as a constraint 
optimisation problem using Integer Linear Programming (ILP) was presented. 
Furthermore, to describe 
abstract 
scenes, Ortiz et al.,~\cite{ortiz2015learning} introduced a machine 
translation-based model 
which translates VDRs to textual descriptions.

\section{KENGIC}
KENGIC is designed to connect visual keywords using $n$-gram graphs by linking 
keywords through intermediary $n$-grams. This graph is then traversed to search 
for paths 
which visit the given keywords. Nodes visited during graph walks are considered 
as phrases for candidate captions. Relevant captions are then
selected based on a cost function (refer to Section~\ref{sec:graph_traversal}) 
which takes the following into consideration: 
(a) the fluency of captions as measured by how probable the sequences of words 
are, (b) the length of captions, (c) the number of keywords found in the 
generated captions, and (d) the number of nouns which have been mentioned 
but not found in the given keywords set.
The high-level architecture (refer to Fig.~\ref{fig:kengic}) is split into two 
modules:

\textbf{Vision}: This module is responsible for the extraction of a 
set of keywords ($\mathcal{K}$) that are relevant to the query image 
($I$). This 
set serves the basis for the generation of a knowledge graph  
($G_{\mathcal{I,\mathcal{K}}}$) which corresponds to image $I$ based on 
keywords 
$\mathcal{K}$. Keywords that are 
grounded in images can be 
detected by either individually trained visual detectors, by scene 
graph generators trained to predict grounded scene graphs, or by 
multi-label models designed to predict image labels including nouns, attributes 
and verbs.

\textbf{Language}: This module, which is the core contribution of 
this 
work, 
handles the generation of knowledge 
graphs by probabilistically linking keywords 
$\mathcal{K}$ of image $I$ through $n$-grams as found in a text corpus 
$T$. This module is designed to 
traverse 
the graph to find the most relevant caption that best describes the 
image based on the given keywords.


\subsection{N-Gram Graph}
In NLP, an $n$-gram refers to sequences of words (or characters) containing $n$ 
elements as found in a sentence. For example, the bigrams of the phrase 
``\textit{a person on  a boat}'' are \{``\textit{a person}'', ``\textit{person 
on}'', ``\textit{on a}'', ``\textit{a boat}''\}. On the other hand, an $n$-gram 
graph is a graph which connects $n$-grams, initially proposed 
in~\cite{giannakopoulos2008summarization} 
as a summarisation method. This was intended to associate pairs of $n$-grams 
with edges to denote how closely each pair is related.
This data-structure was later applied in sentiment 
analysis~\cite{alisopos2011sentiment}, language 
identification~\cite{tromp2011graph} and even in molecular 
representation~\cite{liu2019ngram}. To our knowledge, this is the first time 
that $n$-gram graphs are used in image caption generation. Formally, an 
$n$-gram graph is a graph 
$G^{n}=\{V,E,L\}$, where $V$ is the set of 
vertices consisting of phrases extracted from $n$-grams, $E$ is the set of 
directed edges 
which 
connect phrases represented by vertices $(v_{1}, v_{2})$, and $L$ is a 
function that assigns a label 
to each 
vertex $v_{i}$ after combining and filtering out overlapping $n$-grams. The 
vertices of the $n$-gram graphs are connected based on whether the 
last token of each 
vertex ($v^{-1})$ overlaps with the first token ($v^{0}$) of the remaining 
vertices in $V$. For 
instance, $G^{n=2}$ for the phrase 
``\textit{a person on a boat}'' is defined as follows:
\begin{align*}
	V&=\{\text{``\textit{a person}''}, \text{``\textit{on a}''}, 
	\text{``\textit{boat}''}\},\\
	E&=\{\{\text{``\textit{a person}''}, \text{``\textit{on a}''}\}, 
	\{\text{``\textit{on a}''}, 
	\text{``\textit{boat}''}\}
\end{align*}

\subsection{Graph Generation}
\begin{figure}[h]
	\begin{center}
		\includegraphics[width=0.7\linewidth]{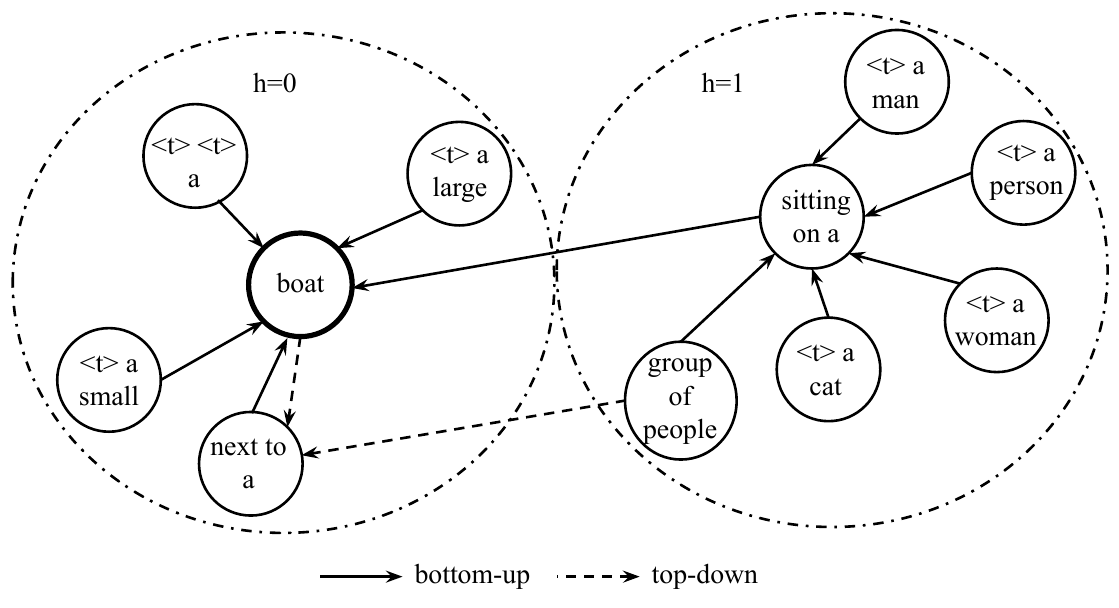}
	\end{center}
	\caption{$N$-gram graph starting from the keyword ``boat''.}
	\label{fig:ngram_graph}
\end{figure}

Given a set of visual keywords, the language module constructs an $n$-gram 
graph in a bottom-up and top-down approach based on padded $n$-grams extracted 
from a given text corpus $T$. For each keyword $w$, the top $k$ frequent 
$n$-grams that end with the word $w$ are considered as parents $\mathcal{P}$ 
of $w$, in such a way that 
$n\text{-gram}^{0:n-1}$ is connected 
to keyword $w$. This is repeated for $h$ hops, where each parent 
$p\in\mathcal{P}~|~p^{0} \neq 
\langle t \rangle$ is connected to its 
$p^{h+1}$ ancestors in a bottom-up approach as illustrated in 
Fig.~\ref{fig:ngram_graph}. For instance, the corresponding five 
topmost $4$-gram parents at $h=0$ for the keyword ``\textit{boat}'' (i.e., 
$\mathcal{P}_{w=``boat''}^{h=0})$ could possibly be: \{``\textlangle 
t\textrangle~\textlangle t\textrangle ~a boat'', ``\textlangle t\textrangle~a 
small boat'', ``\textlangle t \textrangle~a large boat'', ``sitting on a 
boat'', ``next to a boat''\}.
These five $4$-gram sequences are then connected with 
the most probable $n$-gram parents which have their $n^{th}$ word identical to 
the first word in the $n$-grams found in set $\mathcal{P}^{h}_{w}$. This is 
repeated for each keyword in set $\mathcal{K}$ up to a specified number of 
hops 
($h$). As an example, the next hop ($h=1)$  for the sequence ``\textit{sitting 
on a boat}'' connects the most probable $4$-grams that end with 
the word ``\textit{sitting}'' as shown in Fig.~\ref{fig:ngram_graph}.

Once the top-level parents are reached at hop=$h$, the graph 
nodes 
are connected in a 
top-down approach. All unconnected 
nodes that form relevant phrases are connected. This is 
handled by taking into consideration the frequency count of such connections 
and if they are found more than $x$ times in the text corpus, the 
corresponding vertices are connected. For instance, both vertices with labels: 
``\textit{boat}'' and ``\textit{group of 
	people}'' can be linked to the vertex having the phrase ``\textit{next to 
	a}'', 
given that both combined phrases ``\textit{boat next to a}'' and 
``\textit{group of 
	people next to a}'' occur at least $x$ times in the  
text corpus as illustrated by the dotted edges of Fig.~\ref{fig:ngram_graph}. 
This 
constraint was added to reduce rarely occurring connections in the graph 
generation, as well as to 
reduce the graph complexity.

\subsection{Graph Traversal}
\label{sec:graph_traversal}
An image $I$ is described by traversing the corresponding generated $n$-gram 
graph ($G_{I,\mathcal{K}}^{n}$) to search for the most relevant caption that 
best mentions keywords $\mathcal{K}$. The 
search is carried out in a breadth-first 
approach by keeping a list of paths $\mathcal{Q}$ considered as 
relevant captions. The search starts by initialising the list of paths with the 
set of keywords $\mathcal{K}$ (i.e., $\mathcal{Q}=\mathcal{K}$). Each child 
$c$ of the last vertex of $q \in \mathcal{Q}$ is appended with $q$ to 
form path $q + c$. Path $q$ is removed from set $\mathcal{Q}$ while all 
appended paths are added to the set for future concatenation. To reduce the 
time complexity of the graph traversal, the search process considers a total of 
$q_{n}$ paths, while $\mathcal{Q}$ is always populated with the top $y$ best 
paths based on one of the following cost functions 
($\text{fn}_{i}~|~1 \leq i \leq 4$):

1. \textbf{F} = 
$\sum_{i=1}^{}\log P(n_{2}{\text{-gram}}_i)$:
This is used to compute the fluency (F) by calculating the total log 
probability based on each $i^{th}$ $n_{2}$-gram according to the text corpus 
$T$ in order to favour frequently used phrases.

2. \textbf{F + M} = $\sum_{i=1}^{}\frac{\log P(n_{2}{\text{-gram}}_i)}{m}$:
This takes 
into 
consideration both the fluency and matched keywords (M) by taking into 
consideration the number of mentioned keywords $m$ in a given caption. This is 
to penalise captions that do not mention the query keywords.

3. \textbf{F + M + L} = $\sum_{i=1}^{}\frac{\log P(n_{2}{\text{-gram}}_i)}{m 
\times l}$:
This is used to favour fluent and long captions with the maximum number of 
mentioned keywords. Long captions were favoured since fluent captions tend to 
be short captions.

4. \textbf{F + M + L + $N$} =
$\sum_{i=1}^{}\frac{\log P(n_{2}{\text{-gram}}_i)) 
	\times N}{m \times l}$: \\
This favours captions with the highest 
fluency, number of mentioned keywords $m$, caption length $l$ and captions 
which 
have the lowest number of extra nouns $N$ mentioned which are not found in the 
given keywords set.

\section{Dataset}
In this study, the large-scale and most widely used COCO (Common Objects in 
Context) dataset~\cite{lin2014mscoco}, released in 
2014, was used. This contains a 
total of $164,062$ images each captioned with five or six human authored 
captions. This dataset was officially split into training ($82,783$), 
validation 
($40,504$) and 
testing ($40,775$) sets. However, since the captions of the test images are not 
publicly 
available and for consistency with previous studies, the common 
third-party split of Karpathy and Fei-Fei~\cite{karpathy2017deep} was adopted, 
where the original validation 
set was further split 
into validation 
($5,000$) and testing ($5,000$) sets. The remaining images were 
added to the original training set which led to a total of $113,287$ images. 
The human captions found in the training set were used for the 
extraction of $n$-grams needed to generate each $n$-gram graph per query image. 
On 
the other hand, the testing images and their corresponding ground-truth 
captions were used for 
the analyses of this study.

\section{Metrics}
The most popular evaluation metrics were used to measure the 
quality of the generated captions. The 
BLEU~\cite{papineni2002bleu}, ROUGE~\cite{lin2004rouge} and 
METEOR~\cite{banerjee2005meteor} metrics had been adopted from machine 
translation 
and 
document summarisation, while CIDEr~\cite{vedantam2015cider} and 
SPICE~\cite{anderson2016spice} metrics were 
later proposed for image captioning. All metrics, except SPICE, measure the 
$n$-gram overlap between the generated and ground-truth captions. While BLEU 
measures the $n$-gram precision and ROUGE considers the $n$-gram recall, METEOR 
takes into account the precision, recall and synonyms. CIDER 
makes use of TF-IDF to weight $n$-grams and calculates cosine similarity 
between captions. On the other hand, to measure the semantic relatedness which 
$n$-gram based metrics do not consider, SPICE constructs scene graphs of 
reference and candidate captions and compares them based on an F-score computed 
over triplets composed of objects, attributes and relationships.

\section{Experiments}
Two types of keyword sets were used to caption the test images. The first type 
was based on human defined words as extracted from gold captions and these were 
used for 
preliminary studies. This set includes human keywords 
(HK) extracted from the first caption (HC-$0$) as found in the set of 
ground-truth captions and the most frequently used human keywords (HK-f$i$) 
found in all corresponding human captions. On the other hand, the second type 
of 
keywords were based on machine detected words, which include automatically 
detected image objects (Objs) by an off-the-shelf object detector, and image 
keywords detected by a multi-label model (ML).

\subsection{Human Keywords}
The motivation behind these experiments was to analyse the quality of 
the captions that are generated based on keywords used by humans.
To project a sufficient baseline, this study evaluated the quality of the five 
human authored captions (HC-$i$) against the remaining four captions.
Given that no statistical difference was noted between the evaluated human 
captions, the first set of experiments was based on HC-$0$. Keyword 
sets were extracted by using 
the Part-of-Speech (POS) tagger 
based on the Penn Treebank tagset~\cite{marcus1993building} of the Natural 
Language Toolkit 
(NLTK)~\cite{bird2006nltk} library after tokenizing the 
captions. The generated 
captions were evaluated on the remaining ground-truth captions (i.e., 
HC-$\{1-4\}$). The human 
keyword sets which were composed of nouns, attributes, prepositions and verbs 
were used in a composite and non-composite way. The composite sets 
included phrases composed of grouped keywords such as an attribute 
(``\textit{large}'') followed by a
noun (``\textit{boat}'') or  a noun followed by a verb such as 
(``\textit{boat navigating}''). Composite
keywords were used to restrict the model by constraining it to mention such 
keywords in that specified order without leaving any room for discontinuity 
between keywords during the graph generation and path traversal. Furthermore, 
this experiment also sheds light on whether the generation of composite 
keywords 
would improve the quality of the generated captions. These 
experiments were split into the following six categories to reflect the 
combinations of nouns, attributes, prepositions, and verbs:

\textbf{HK-n}: Human keywords consisting of nouns only (e.g., 
``\textit{boat}'', ``\textit{person}'').

\textbf{HK-na}: Human keywords consisting of nouns and attributes 
(e.g., 
``\textit{large}'', ``\textit{boat}'').

\textbf{HK-nap}: Human keywords consisting of nouns, attributes and 
prepositions 
(e.g., ``\textit{large}'', ``\textit{boat}'', `` \textit{near}'').

\textbf{HK-napv}: Human keywords consisting of nouns, attributes, 
prepositions and verbs (e.g., ``\textit{large}'', ``\textit{boat}'', `` 
\textit{near}'', ``\textit{navigating}'').

\textbf{HK-(na)}: Composite human keywords composed of an attribute 
followed by a noun (e.g., ``\textit{large boat}'').

\textbf{HK-(nv)}: Composite human keywords composed of a noun 
followed by a verb (e.g., ``\textit{boat navigating}'').

Rather than using
human extracted keywords from one single caption, another experiment
was set to examine the quality of the generated captions based on
the most salient keywords that humans choose when describing
images. This was carried out by selecting the most commonly used
keywords found in the set of corresponding ground-truth captions
per image. This was handled by considering keyword sets with cumulative 
frequency count per word. Each set was denoted by HK-f$i$, where $i$ corresponds
to the minimum frequency count of each word. For example, HK-f$2$ consists of 
keywords which occur at least two times in the set of ground 
truth 
captions, while HK-f$5$ consists of keywords which are common in 
all five ground-truth captions. For consistency with HK, the generated captions 
were compared against HC-$\{1-4\}$. Captions 
simply composed from frequent keywords (HK-f$i$(kw)) were also evaluated to 
assess their effect on the evaluation metrics.

\subsection{Detected Keywords }
The third experiment was set to investigate the quality of the generated 
captions based on automatically detected keywords. Since captions based on the 
human-authored keywords were compared against four captions, in 
this experiment, the generated captions were evaluated on both four and five 
reference captions. For this purpose, both objects (Objs) and multi-label 
keywords (ML) keywords were predicted as follows:

\textbf{Objs}: Objects were detected using a pre-trained Faster 
R-CNN~\cite{ren2015faster} capable of detecting up to $80$ COCO 
object classes. The average 
number of detected objects per image was $3.4$. Given 
that $15$ out of the $80$ object classes consisted of two words, an additional 
experiment was performed to split 
(+sp) keywords. Multiple objects (+multi) were also considered when multiple 
objects having the 
same class were detected.

\textbf{ML}: A multi-label model trained to predict multi labels (ML) relevant 
to query 
images was used. Having a 
model trained to predict labels that include nouns, attributes and verbs 
should provide a set of rich keywords. For 
this reason, the ML-Decoder~\cite{ridnik2021ml} model which follows a 
Transformer-based encoder-decoder pipeline was used. A pre-trained ML-Decoder 
was adopted to detect the $80$ COCO object labels (ML-objs) to provide 
comparison with the Faster R-CNN object detector results. An ML-Decoder 
pre-trained on OpenImages~\cite{kuznetsova2020open} was also fine-tuned to 
predict keywords extracted from COCO image captions using the same 
hyperparameters reported by Ridnik et al.~\cite{ridnik2021ml} for COCO dataset. 
The fine-tuning was performed on the vocabularies $\mathcal{V}$ extracted from 
the captions found in COCO Karpathy's train set as follows:

\textbf{Cleaned-$w$ (C-$w$)}: The top $w$ frequent words which are 
not considered as stop words and which do not contain any numbers in the 
human captions.

\textbf{Cleaned + Lemmatised-$w$ (CL-$w$)}: The top $w$ frequent 
cleaned and 
lemmatised (L)
words found in the human captions. This is used to reduce the complexity 
of the used vocabulary set by using base words which are commonly referred 
to as lemmas.

\textbf{Cleaned + Lemmatised + POS Filtering (CLP-$w$)}: Consists of the top 
$w$ 
cleaned, lemmatised and POS filtered words. This is used to further reduce the 
vocabulary set by words consisting of nouns, attributes, and verbs.

These vocabularies $v\in\mathcal{V}$ for $w \in \{1000, 
2000,3000\}$ were used as 
sets for the extraction of keywords $\mathcal{K}$ for query images. To 
predict the salient keywords, the ML-Decoder was 
explicitly trained to predict keywords according to their frequency as found 
in the corresponding ground-truth image captions. Similar to how HK-f${i}$ 
based 
keywords were extracted, the ML-Decoder was 
trained to 
predict labels based on their cumulative frequency (f$j~|~1 \leq j 
\geq 
5 $). For instance, if $j=z$, the ML-Decoder was trained to predict  keywords 
$\mathcal{K}$ which occur at least $z$ times 
across 
all corresponding
captions of a given image and which are all found in the given vocabulary $v$. 
It was observed that the performance of the ML-Decoder peaks at $j=3$ followed 
closely by $j=2$ in all tested scenarios. Similar results were obtained when 
using different vocabularies and sizes. For this reason, ML-C-$w$-f$2$ and 
ML-C-$w$-f3 vocabulary sets were used as they are broader and less restrictive.

\section{Results}
This section reports and discusses the quantitative results of this study. A 
qualitative analyses was also conducted to get deeper insights behind the 
generated captions.

\subsection{Hyperparameter Optimisation}
Hyperparameter optimisation (HPO) was carried out based on the baseline human 
extracted keywords. Since no statistical significance was noted when evaluating 
the quality of the five human ground-truth captions, the optimisation was 
carried 
out by a grid search on HK-n
keywords set extracted from HC-0 by varying the following parameters as 
follows: 
$n \in \{3,4\}$, $n_{2} \in 
\{3,4\}$, $h \in \{1,2\}$, 
$\text{fn}_{i}~|~1 \leq i \leq 4$, 
whilst the values of $k$, $x$, $y$ 
were manually set to $5$ and $q_n$ was set to $150$. This results in 
a total of $32$ configurations. Due to the time 
complexity of high-ordered graphs, the optimisation was carried out on a random 
sample chosen from the validation set. Based on a population size of $5000$, a 
sample size of $357$ estimates the population results with a $95\%$ confidence 
level and $5\%$ margin of error. Therefore, a sample size of $500$ images was 
chosen for the validation process. Since image caption generators are generally 
optimised on the CIDEr metric~\cite{anderson2018bottom, yang2020auto}, the 
hyperparameters ($n=3, h=1, n_{2}=3, \text{fn}_{i}=4$) which maximised the 
CIDEr score were selected.

\begin{table*}
	\begin{center}
		{\small{
				\begin{tabular}{@{}lllllll@{}}
					\toprule
					Keywords  & B-1      & B-4     & M      & R     & C   & 
					S       \\ \midrule
					HC-0      & 63.6        & 19.9       & 24.4        & 
					47.3        & 89.3        & 21.2        \\ \midrule
					HK-n      & 50          & 13.4       & 18.4        & 
					37.6        & 66.4        & 16.3        \\
					HK-na     & 54.3        & 14.4       & 19.3        & 
					\textbf{38}          & \textbf{68.3}        & 
					\textbf{16.8}        \\
					HK-nap    & 54.6        & 14.3       & 19.4        & 
					\textbf{38}          & 67.6        & \textbf{16.8}       \\
					HK-napv   & \textbf{55.5}        & \textbf{15.5}       & 
					\textbf{20.3}        & 37.7        & 65.2        & 
					16.3        \\
					HK-(na)   & 26.1        & 7          & 12.2        & 
					25          & 42.1        & 10.3        \\
					HK-(nv)   & 18.2        & 4.9        & 10.4        & 
					20.7        & 35.3        & 8.3         \\ \midrule
					HK-f1(kw) & 59.3 (\textbf{85.6}) & 19.2 (0)   & 
					\textbf{30.8} (24.8) & 38.9 (23.7) & 25.6 (53)   & 
					\textbf{33.7} (17.6) \\
					HK-f2(kw) & 76.1 (55.4) & \textbf{28.6} (0.9) & 28.2 (24.2) 
					& \textbf{47.6} (32.8) & \textbf{112.3} (91)  & 25.8 (16.1) 
					\\
					HK-f3(kw) & 35.2 (10.8) & 14.9 (0.3) & 20.4 (16.9) & 39.8 
					(27.4) & 91.9 (56)   & 18 (11.7)   \\
					HK-f4(kw) & 4 (0.6)     & 1.8 (0)    & 11.3 (9.7)  & 24.5 
					(18.7) & 50.7 (34.8) & 10.8 (8)    \\
					HK-f5(kw) & 0 (0)       & 0 (0)      & 4.4 (4)     & 10.7 
					(9.5)  & 20.6 (17.5) & 5.1 (4.5)   \\ \bottomrule
				\end{tabular}
		}}
	\end{center}
	\caption{Results metrics of captions generated based on HK keywords 
	alongside the metrics of HC-0 as computed on COCO Karpathy's test split 
	using HC-\{1-4\}. The best results are marked in boldface. The metrics 
	B-$n$, M, R, C, and S correspond to BLUE-$n$, METEOR, ROUGE-L, CIDEr and 
	SPICE respectively.}
	\label{tbl:hk_metrics}
\end{table*}

\begin{table*}
	\begin{center}
		{\small{
				\begin{tabular}{@{}lllllll@{}}
					\toprule
					Keywords      & B-1      & B-4      & M      & R     & 
					C       & S       \\ \midrule
					Objs          & 10.7 (11.6) & 1.8 (2.1)   & 7.8 (8.2)   & 
					17.9 (18.6) & 22.9 (23.7) & 6.5 (5.8)   \\
					Objs+sp       & \textbf{38.1} (\textbf{40.3}) & \textbf{7} 
					(\textbf{7.9})     & \textbf{13.5} (\textbf{14.2}) & 
					\textbf{28.7} (\textbf{29.8}) & \textbf{36.4} 
					(\textbf{37.5}) & \textbf{10.6} (9.6)  \\
					Objs+sp+multi & 34.8 (37)   & 5.9 (6.6)   & 12.6 (13.2) & 
					27.2 (28.4) & 34.7 (35.6) & \textbf{10.6} (\textbf{9.7})  
					\\ \midrule
					ML-objs       & 10.2 (11.2) & 1.7 (2)     & 8.2 (8.7)   & 
					18.6 (19.4) & 24.3 (25.2) & 7.2 (6.4)   \\
					ML-C-1K-f2    & 60.1 (63.6) & 15.7 (18)   & 21 (22)     & 
					\textbf{38.7} (40.3) & \textbf{68.1} (\textbf{69.8}) & 19.6 
					(18.3) \\
					ML-C-2K-f2    & \textbf{62.7} (\textbf{66.3}) & 
					\textbf{16.1} (\textbf{18.6}) & \textbf{21.5 } 
					(\textbf{22.6}) & \textbf{38.7} (40.4) & 66.2 (67.8) & 
					\textbf{19.8} (\textbf{18.5}) \\
					ML-C-3K-f2    & 60.7 (64.3) & 15.8 (18.3) & 21.1 (22.2) & 
					\textbf{38.7} (\textbf{40.5}) & 67.7 (69.4) & 19.7 (18.4) \\
					ML-C-1K-f3    & 32.6 (34.7) & 9.3 (10.9)  & 16 (16.9)   & 
					32.9 (34.4) & 56.5 (58)   & 15.7 (14.1) \\
					ML-C-2K-f3    & 29.2 (31.2) & 8.5 (10)    & 15.5 (16.4) & 
					32.1 (33.6) & 55.1 (56.4) & 15.4 (13.8) \\
					ML-C-3K-f3    & 31.6 (33.8) & 9 (10.5)    & 15.9 (16.7) & 
					32.8 (34.2) & 55.9 (57.4) & 15.8 (14.2)
					\\ \bottomrule
				\end{tabular}
		}}
	\end{center}
	\caption{Results metrics of captions generated based on predicted keywords 
	as computed on COCO Karpathy's test split. Captions were evaluated on four 
	and five (in brackets) captions.}
	\label{tbl:predicted_metrics}
\end{table*}

\begin{table}
	\begin{center}
		{\small{
				\begin{tabular}{@{}lllllll@{}}
					\toprule
					Model & B-1 & B-4 & M & R & C $\downarrow$
					& S \\ \midrule
					\textit{Paired}     &    &   &  &  &  &  \\
					~~Mind's Eye~\cite{chen2015mind}     & -   & 18.8  & 
					19.6 
					& - 
					& - & - \\
					~~NeuralTalk~\cite{karpathy2017deep}     & 62.5   
					& 
					23.0  & 19.5 & - & 66.0 & - 
					\\
					~~Up-Down~\cite{anderson2018bottom}     & 77.2      & 
					36.2  
					& 
					27.0 & 54.9 & 113.5 & 20.3 \\
					~~SGAE-KD~\cite{yang2020auto} & 78.2  & 37.3 & 28.1 & 57.4 
					& 
					117.1 
					& 21.3 \\
					~~MT~\cite{yu2020multimodal} & 77.3  & 37.4 & 28.7 & 57.4 & 
					119.6 
					& - \\
					\midrule
					\textit{Unpaired}        &    &   &  &  &  &  \\
					~~Pivoting~\cite{gu2018unpaired} & 46.2  & 
					5.4 & 13.2 & 
					- 
					& 
					17.7 
					& - \\
					~~Adverserial~\cite{feng2019unsupervised} & 58.9  & 18.6 
					& 17.9 
					& 43.1 & 
					54.9 
					& 11.1 \\
					~~USGAE~\cite{yang2020auto} & 60.8 & 17.1 & 19.1 & 43.8 & 
					55.1 
					& 12.8 \\
					~~Multimodal~\cite{laina2019towards} & -  & 19.3 
					& 20.2 & 
					45.0 & 
					61.8 
					& 12.9 \\
					~~IGGAN~\cite{cao2020interactions} & - & 21.9 & 46.5 & 21.1 
					& 
					64.0 
					& 14.5 \\
					~~\textbf{ML-C-2K-f2} & \textit{66.3}  & 18.6 
					& 
					\textbf{22.6} & 
					40.4 & 
					67.8 
					& \textbf{18.5} \\
					~~Graph-Align~\cite{gu2019unpaired} & \textbf{67.1}  & 
					\textit{21.5} & 
					20.9 & \textit{47.2} 
					& 
					69.5 
					& 15.0 \\
					~~\textbf{ML-C-1K-f2} & 63.6 & 18 & 22.0 & 
					40.3 
					& 
					\textit{69.8} 
					& \textit{18.3} \\
					~~SCS~\cite{ben2022unpaired} & \textbf{67.1}  & 
					\textbf{22.8} & 
					\textit{21.4} & \textbf{47.7} 
					& 
					\textbf{74.7} 
					& 15.1 \\ \bottomrule
				\end{tabular}
		}}
	\end{center}
	\caption{Performance of KENGIC based on \textnormal{ML-C-1/2K-f2} keyword 
	sets 
	against 
	state-of-the-art image caption generators on COCO Karpathy's test split.}
	\label{tbl:sota_comparison}
\end{table}

\subsection{Quantitative Analysis}

The evaluation of the captions generated by KENGIC based on human 
extracted keywords is presented in 
Table~\ref{tbl:hk_metrics}. The first section of the table consists of captions 
which were generated based on keywords extracted from one human authored 
caption (i.e., HC-0), whilst the second section presents the evaluation of the 
captions generated based on the salient words found across all ground-truth 
captions (i.e., HK-f$i$). It was clear that the captions generated based on the 
non-composite 
keywords set are highly comparable and significantly better than those produced 
by the composite keywords. Although not very significant, HK-napv obtained the 
highest scores, except on ROUGE-L, CIDEr and SPICE which peaked when using 
HK-na. On the other hand, the composite keyword sets  
were found to restrict the model and led to the lowest scored 
captions.

When considering the best 
performing set of salient keywords (i.e., HK-f2), the metrics improved 
substantially over HK across all metrics, especially in CIDEr score which 
increased from $68.3$ to $112.3$ when compared to HK-na. Surprisingly, HK-f$1$ 
obtained the highest SPICE score of $33.7$ despite using an average of $18.09$ 
keywords per image. The 
evaluation metrics 
revealed that captions composed of only frequent keywords (kw) obtained 
considerably 
high scores. In fact, 
HK-f2(kw) even exceeded the quality of HC-0 in terms of CIDEr. These results 
confirm that metrics give an important weight to the mentioned keywords and pay 
less attention to the sentence structure and the order of the used words. This 
observation raises important questions on how 
captions are being evaluated by the current popular metrics.

The generated captions based on the predicted keyword sets were compared 
against five and four captions to provide fair comparison with the results 
generated based on the human extracted keywords as tabulated 
in Table~\ref{tbl:predicted_metrics}. When compared to HK-f2, the 
quality of the generated captions when using the predicted Objs keywords 
set decreased 
substantially due 
to the limited vocabulary set size of the used detector (i.e., $80$ 
objects). Considerable improvements were 
recorded with the introduction of the two-word splitting (Objs+sp). As 
previously 
confirmed with 
composite human extracted keywords (i.e., HK-(na), HK-(nv)), this shows 
that multi-word keywords constrain KENGIC. Objs+sp+multi ended up being less 
effective in all metrics, 
except in SPICE score where no changes were recorded.
The quality of the generated captions based on the ML-objs keywords set was 
found similar to that obtained when using the Objs 
keyword set. This 
confirmed the level of comparability between the outputs of the Faster 
R-CNN~\cite{ren2015faster} and ML-Decoder~\cite{ridnik2021ml} when both are 
trained to detect objects in images. From the results listed in 
Table~\ref{tbl:predicted_metrics}, it was found that ML-C-2K-f2 scored 
best on all BLEU, METEOR and SPICE scores, whilst ML-CL-1K-f2 and ML-C-1K-f2 
recorded a slight improvement on ROUGE-L and CIDEr respectively. Similar to the 
results based on HK-f$i$, predicted keywords based on a frequency of $2$ led 
to better results. Similar observations were noted when 
evaluating the captions against the full set of ground-truth captions. As 
expected, a slight improvement was observed when compared to the results 
evaluated on four captions, except in SPICE score, where a slight decrease was 
noted in all configurations.

\begin{figure}[h]
	\begin{center}
		\includegraphics[width=\linewidth]{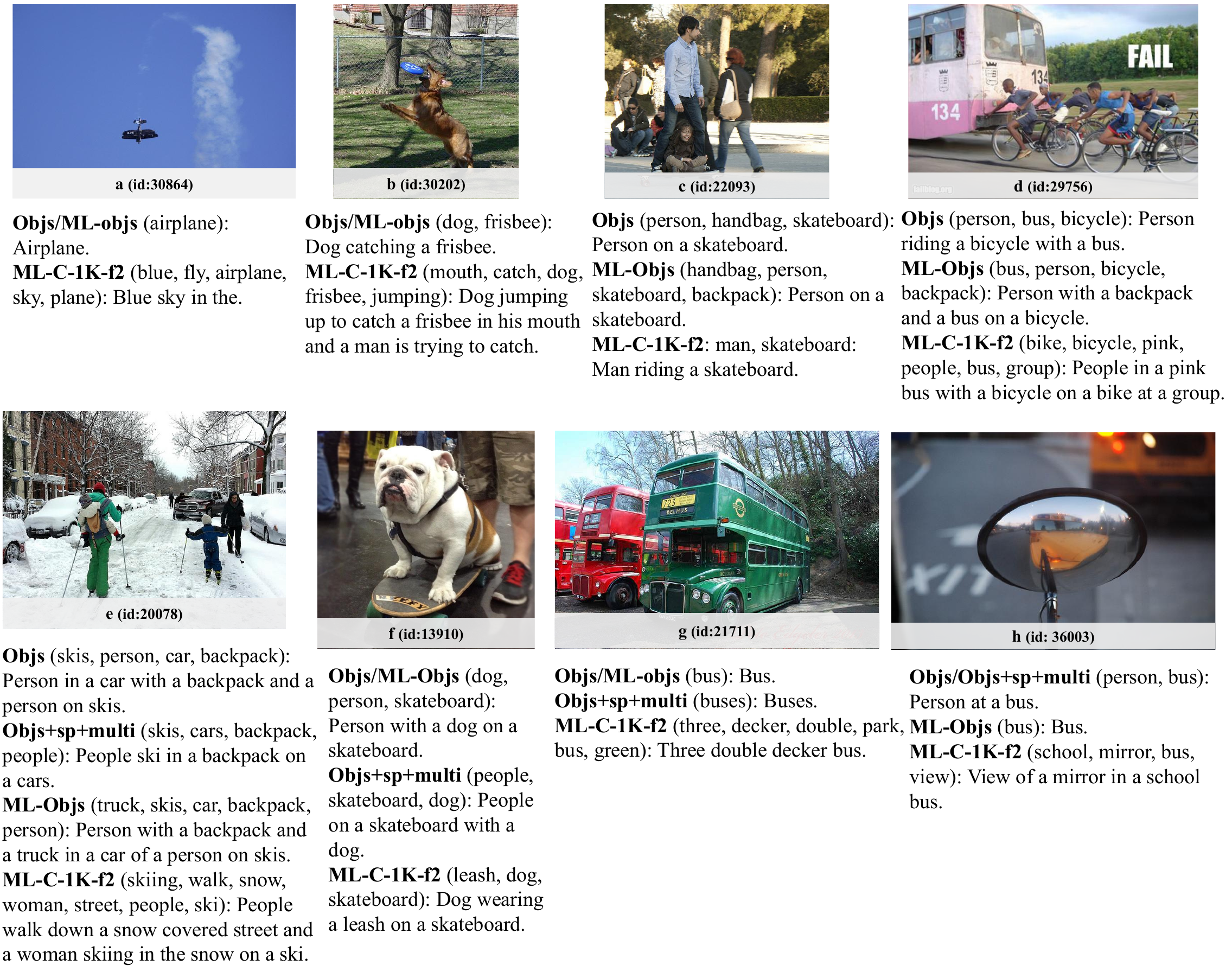}
	\end{center}
	\caption{Examples of low quality (top row) and good quality (bottom row) of 
	captions generated by KENGIC based on predicted keywords.}
	\label{fig:images}
\end{figure}

\subsection{Comparison with State-of-the-Art Methods}
The results generated by KENGIC based on ML-C-1K-f2 and ML-C-2K-f2 were 
juxtaposed with current state-of-the-art and benchmark image caption generators 
in Table~\ref{tbl:sota_comparison}. The results are grouped in two sections to 
distinguish between models which are trained in the paired and unpaired 
setting. As 
expected, KENGIC falls short when compared with models trained in the paired 
setting 
since 
it does not make use of any end-to-end training on image-caption pairs. 
Therefore, 
for fair comparisons, KENGIC was compared with unpaired 
image caption generators. It was found that KENGIC performance is very close to 
that of state-of-the-art unpaired generators, and in some metrics, it even 
surpasses current benchmark models. Overall, the model was found to be on par 
with the current top two best performing 
unpaired models.
Despite its simplicity and the fact that the 
vision and language domains are not connected using an end-to-end pipeline, 
KENGIC 
based on ML-C-2K-f2 keywords set achieved the highest METEOR and SPICE scores 
of $22.6$ and $18.5$ 
respectively. On the other hand, the use of ML-C-1K-f2 keywords ranked the 
model second in terms of CIDER ($69.8$) when compared to the more complex 
SCS~\cite{ben2022unpaired} model that is based on adversarial training.

\subsection{Qualitative Analysis}

To complement the quantitative analysis, a 
qualitative analysis on $200$ randomly sampled images was conducted to get a 
deeper insight into the workings of the system. It was found 
that in simple scenarios KENGIC was unable to construct meaningful and relevant 
captions (Fig.~\ref{fig:images}(a)). The model was also found to hallucinate 
most probably because KENGIC prefers longer captions 
(Fig.~\ref{fig:images}(b)). Furthermore, the synonymous nature of the predicted 
keywords proved to be a problem for the overall generation process. As shown 
in~Fig.~\ref{fig:images}(a), the 
keywords ``airplane'' and ``plane'' of ML-C-1K-f2 probably confused the 
generator and led to a caption which does not mention the main subject of the 
image. Some captions also lacked important details. For 
example, all the captions for Fig.~\ref{fig:images}(c) lacked 
mentioning the girl that is sitting under the man. Images were also captioned 
incorrectly 
due to the complexity and the degree of reasoning they require (e.g., 
Fig.~\ref{fig:images}(d)).
On the other hand, KENGIC was also found to generate accurate captions, even in 
less 
common scenarios as shown in the bottom row of 
Fig.~\ref{fig:images}. Despite the relevancy of such  captions, the 
corresponding metrics scores were found to be relatively low. This confirmed 
the lack of correlation between caption quality and the metrics, and the need 
for more robust evaluation metrics. Surprisingly, this analysis also revealed 
that the ML-Decoder was able to implicitly learn how to count objects in 
images as shown in Fig~\ref{fig:images}(g). KENGIC was also found 
robust in complex scenarios like in Fig~\ref{fig:images}(h).

\section{Conclusions and Future Work}
In this paper, a keyword driven and $n$-gram graph based image caption 
generator was proposed as an alternative approach in automatic image 
captioning. This was proposed for better grounding of captions using a more 
explainable approach, whilst reducing the dependency of large-scale paired 
image-caption datasets. Results confirmed the possibility of generating high 
quality captions from keywords through $n$-gram graphs. Both quantitative and 
qualitative analysis showed that the model can effectively introduce additional 
words alongside the given keywords, as well as construct fluent and succinct 
captions. This study also concludes that metric scores were generally penalised 
with the introduction of prepositions and verbs. This could either be because 
these words where missing or used sparingly in the gold standard captions; or 
else because they were not part of the used text corpus and therefore limited 
the $n$-gram graph construction. As expected, this study also confirmed that 
captions which use words that are frequently used in ground-truth captions 
benefit from higher scores. This is not surprising since evaluation metrics 
generally measure the maximum overlapping $n$-gram sequences found between the 
candidate and reference captions (BLEU, METEOR and ROUGE-L), whilst CIDEr and 
SPICE consider all captions in their final metric and therefore benefit from 
commonly used words across captions. KENGIC based on predicted visual keywords 
showed promising results when compared with current 
state-of-the-art image caption generators trained in an unpaired setting. In 
the future, a complexity analysis will be conducted to improve the 
runtime performance of the system. KENGIC will also be studied in an 
unsupervised way 
where images and the 
used text corpus will be unrelated. Furthermore, this approach will also be 
explored in Visual Question Answering (VQA).

\bibliographystyle{IEEEtran}
\bibliography{bibliography}


\end{document}